# A Survey From Distributed Machine Learning to Distributed Deep Learning

Mohammad Dehghani[1*], Zahra Yazdanparast[2]

[1] School of Electrical and Computer Engineering, University of Tehran, Tehran, Iran,

[2] School of Electrical and Computer Engineering, Tarbiat Modares University, Tehran, Iran,

## Abstract

Artificial intelligence has made remarkable progress in handling complex tasks, thanks to advances in hardware acceleration and machine learning algorithms. However, to acquire more accurate outcomes and solve more complex issues, algorithms should be trained with more data. Processing this huge amount of data could be time-consuming and require a great deal of computation. To address these issues, distributed machine learning has been proposed, which involves distributing the data and algorithm across several machines. There has been considerable effort put into developing distributed machine learning algorithms, and different methods have been proposed so far. We divide these algorithms in classification and clustering (traditional machine learning), deep learning and deep reinforcement learning groups. Distributed deep learning has gained more attention in recent years and most of the studies have focused on this approach. Therefore, we mostly concentrate on this category. Based on the investigation of the mentioned algorithms, we highlighted the limitations that should be addressed in future research.

**Keywords:** Artificial intelligence, Machine learning, Distributed machine learning, Distributed deep learning, Ditributed reinforcement learning, Data-parallelism, Model-parallelism.

## Introduction

Artificial intelligence (AI) is a rapidly developing field that uses knowledge to simulate human behaviors (1) and train computers to learn, make judgments, and make decisions similarly to humans (2, 3). In other words, AI involves developing techniques and algorithms that are capable of thinking, acting, and implementing tasks using protocols that are otherwise beyond human comprehension (4).

Machine learning (ML) is a subset of AI that learns from historical data, without being explicitly programmed (5). ML algorithms can be used to analyze data and build data-driven systems, including classification, clustering, regression, association rule mining, and reinforcement learning (6, 7). Deep learning is a branch of machine learning that uses artificial neural networks to intelligently analyze large amounts of data (8, 9). The digital world has been provided with a large amount of data in many different areas, such as cybersecurity, business, health, and the Internet of Things (IoT). This wealth of data is analyzed by machine learning, one of the most popular technologies in Industry 4.0, in order to develop smarter and more automated systems (10). ML algorithms have been widely used in many application domains, such as computer vision (11, 12) and natural language processing (NLP) (13). They have proven to be effective across a wide range of industries comprising education (14), healthcare (15), marketing (16), transportation (17), energy (18), combustion science (19), and manufacturing (20).

A machine learning solution is influenced by the characteristics of the data and the performance of the learning algorithms (10). Traditionally, a bottleneck in developing more intelligent systems was data availability which is no longer the case. However, the problem now is that learning algorithms are unable to employ all the data within a reasonable period of time for learning. Creating an effective ML model is generally a complex and time-consuming process that involves selecting an appropriate algorithm and developing an optimal model architecture (21).

In order to train ML models over large volumes of data, one machine's storage and computation capabilities are insufficient. One solution for this challenge is employing distributed machine learning for the executation of ML programs on clusters, data centers, and cloud providers (22). It can divide the learning process across several workstations, achieving scalability of learning algorithms (23). Generally, distributed machine learning allows a cloud or server to collect combined models from multiple participants, with each participant training their own model locally (24). Furthermore, it allows the handling of naturally distributed data sets, which is a common scenario in many real-world applications (23). Many researchers are currently focusing on algorithmic correctness and faster convergence rates of ML models (25, 26).

This paper presents a comprehensive view of various types of distributed machine learning algorithms to overcome the problems of traditional machine learning approaches. The main contribution of this study is the introduction of different distributed machine learning techniques, which can provide a valuable guide for those in academia and industry interested in studying, researching, and developing machine learning-based data-driven automation and intelligent systems.

Compared to traditional machine learning algorithms, distributed deep learning has received more attention in studies. In recent years, deep learning has achieved tremendous success in various areas, such as image processing, NLP, and speech recognition. One of the reasons for this success is the availability of a large amount of data and the increased size of the deep learning model. As deep learning continues to improve, increasing its scalability is essential. Currently, distributed deep learning is gaining increasing recognition to overcome the challenges of deep learning. The benefits of distributed deep learning include (27):

- Increased scalability: Distributed deep learning is increasingly necessary as neural networks and datasets grow. The training process could be scaled to handle more extensive datasets and models, and more computers can be added to the distributed system to increase scalability and enable quicker training cycles.

- Resource utilization: By spreading the work across several computers, we may use the already available resources and reach a higher level of parallelism, resulting in shorter training durations and less resource usage.

The rest of the paper is organized as follows. Section 2 provides an overview of differences between this survey and existing surveys. Section 3 provides a taxonomy of distributed machine learning algorithms. Section 4 concludes the survey and considers the direction of future research.

## Related Works

To the best of the authors' knowledge, very few works have proposed to survey distributed algorithms. This article presents a literature review on distributed machine learning algorithms and states their distinctions from existing surveys. This study provides an extensive overview of the current state-of-the-art in the field by outlining the challenges and opportunities of distributed machine learning over conventional (centralized) machine learning and discussing the techniques used for distributed machine learning.

Verbraeken et al. (28) discussed distributed techniques and systems and covered various aspects of distributed machine learning including its challenges and opportunities. Langer et al. (29) investigated the fundamental principles for training deep neural network (DNN) in a cluster of independent machines. Moreover, they analyzed the common attributes of training deep learning models and surveyed the distribution of such workloads in a cluster to attain collaborative model training. Ouyang et al. (30) provided a comprehensive survey of communication strategies for distributed DNN. They considered algorithm and network optimizations to diminish the communication overhead in distributed DNN training. Tang et al. (31) have detailed discussions on communication compression techniques.

Xing et al. (22) investigated different synchronization schemes, scheduling, workload balancing schemes, and communication types. Nassef et al. (32) overviewed distributed machine learning architectures for 5G networks and beyond. They also considered optimizing communication, computation, and resource distribution to improve the performance of distributed machine learning in 5G networks. Mayer et al. (33) overviewed deep learning infrastructures, parallel deep learning methods, scheduling, and data management. Muscinelli et al. (34) surveyed the influential research of distributed learning technologies playing a critical role in the 6G networks. In particular, they reviewed federated learning and multi-agent reinforcement learning algorithms. They discussed several emerging applications and their fundamental concepts relating to their implementation and optimization.

An overview of distributed deep reinforcement learning is provided by Yin et al. (35). Their study compared classical distributed deep reinforcement learning (DRL) methods and examined important factors that contribute to efficient distributed learning. Furthermore, they discuss both single player, single agent distributed DRLs as well as multiple players, multiple agents. A further review was carried out of recently released toolboxes which assist in the realization of distributed

DRL. Antwi-Boasiako et al. (36) provided a comprehensive survey of privacy issues in distributed deep learning. They described various cryptographic algorithms and other techniques that can be used to preserve privacy, as well as their benefits and drawbacks.

Other surveys lack detailed discussions on the distribution of machine learning algorithms, which are introduced explicitly in this survey. We cover a variety of algorithms of traditional machine learning (classification and clustering algorithms), deep learning, and reinforcement learning.

## Distributed machine learning algorithms

A number of popular algorithms for distributed machine learning are described in this section. Fig. 1 represents a classification of these algorithms.

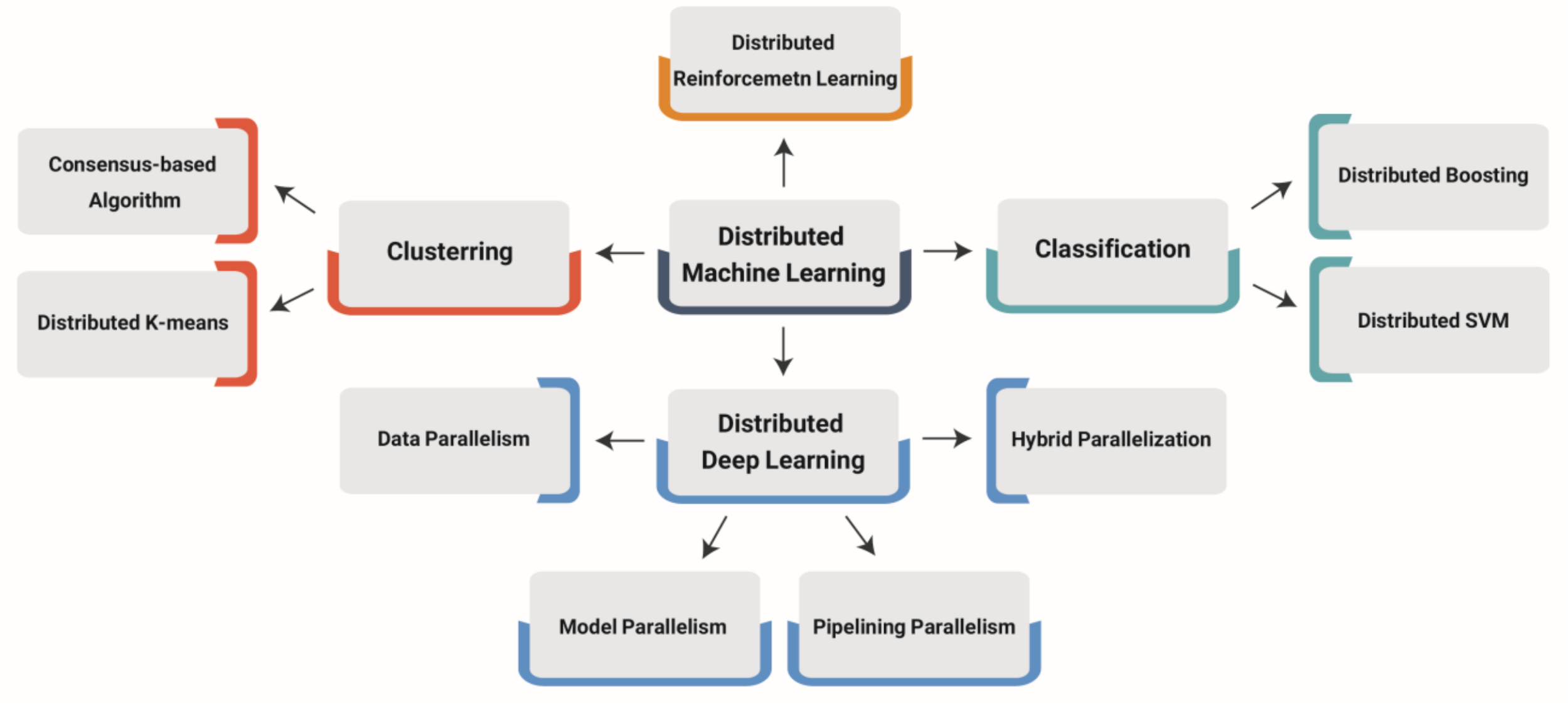

**Fig. 1** Ditributed machin learning algorithms.

### Distributed classification

Classification is a supervised method in machine learning that uses label data to predict the future. A classification model trains using historical data to produce a function that can predict outputs for new unseen data. Classification outputs are categorical (37). Two popular distributed classification methods are boosting and support vector machines (SVM), which we discuss in the following. These algorithms are summarized in Table 1.

#### *Distributed Boosting*

Boosting is a technique used in machine learning that trains an ensemble of so-called weak learners to produce an accurate model, or strong learner. It relies on the idea that learning and combining many weak classifiers instead of learning a single strong classifier can achieve better performance (38).

In distributed boosting, AdaBoost (39) is modified for use in distributed environments (23). Lazarevic and Obradovic (40) proposed a framework for the integration of specialized classifiers learned from very large, distributed datasets that cannot be stored in the main memory of a computer. The suggested method combines classifiers from all sites and creates a classifier ensemble on each site. Filmus et al. (41) proposed a distributed boosting algorithm that is resilient to a limited amount of noise using Impagliazzo's hard-core lemma (42). This algorithm is novel because it applies non-standard boosting that identifies small "hard" sets so that any hypothesis derived from the class will have high error rates.

Sarnovsky and Vronc (43) implemented a distributed boosting algorithm based on MapReduce and employed the GridGain framework for distributed data processing. The algorithm works as follows: (i) The master node computes the dataset statistics, including the number of categories and the distribution of indexed terms within the dataset. (ii) The number of subtasks (jobs) necessary to construct the final model is calculated based on the gridSize parameter. (iii) AdaBoost is implemented to train the model.

Cooper et al. (44) presented two distributed boosting algorithms. The first algorithm uses the entire dataset to train a classifier and requires significant communication between the distributed sites. In contrast, the second algorithm requires little communication but trains its final classifier using a subset of the dataset.

***Distributed SVM***

SVM is a linear binary classifier that identifies a single boundary between classes. It seeks to overcome the local extremum dilemma inherent in other machine learning techniques by solving a convex quadratic optimization problem (45). SVM determines an optimal hyperplane (a line in the simplest case) by which it is possible to divide the dataset into a discrete number of classes. To optimize the separation, SVM employs a portion of the training sample that lies closest to the optimal decision boundary in the feature space (46).

SVM training requires quadratic computation time. To its speed up, several distributed computing paradigms have been investigated by dividing the training dataset into smaller sections and processing each section in the parallel cluster of computers (47). Lu et al. (48) proposed a distributed parallel support vector machine (DPSVM) that exchanges support vectors among a network of strongly connected servers. It results in limited communication costs and fast training times for multiple servers working concurrently on distributed datasets. DPSVM uses SVM as a local classification mechanism for subsets of training data within a strongly connected network.

According to Alham et al. (49), SVM training is computationally intensive, particularly with large datasets. To address this issue, they presented MRSMO, a distributed SVM algorithm based on MapReduce for automatic image annotation. MRSM partitions datasets into small subsets and optimizes them across a cluster of computers. Ke et al. (50) proposed a method for distributed SVM, where the local SVMs use the state-of-the-art SVM solvers and implement it on MapReduce to shorten the communication between nodes. Wang et al. (51) described a spatially distributed SVM method for estimating shallow water bathymetry from optical satellite imagery. This method uses SVMs that have been trained locally and spatially weighted votes to make predictions. According to the results, the localized model reduced the RMSE by 60%.

**Table 1** Distributed classification.

| **Algorithm** | **Articles** | **Year** | **No of references** | **Simulation/ Dataset** | **Evaluation metrics** |
|---|---|---|---|---|---|
| Distributed Boosting | (41) | 2022 | 26 | - | • Accuracy<br>• Correctness<br>• Communication complexity |
| | (44) | 2017 | 20 | • ocr17<br>• ocr49<br>• forestcover12<br>• particle<br>• ringnorm<br>• twonorm<br>• Yahoo! | • Error |
| | (43) | 2014 | 19 | • Reuters-21578<br>• Medlin | • Time |
| | (40) | 2002 | 31 | • Covertype<br>• Pen-based digits<br>• Waveform<br>• LED | • Accuracy<br>• Speedup |
| Distributed SVM | (51) | 2019 | 32 | • Optical satellite images | • RMSE |
| | (50) | 2015 | 14 | • Spiral data set<br>• MNIST<br>• COVERTYPE | • Integrations<br>• Parallel Speed-up |
| | (49) | 2011 | 40 | • Images from Corel database | • accuracy<br>• training times |
| | (48) | 2008 | 17 | • MNIST | • CPU seconds<br>• Number of iterations<br>• Communication overhead |
| | (47) | 2003 | 16 | • Handwritten Chinese database ETL9B | • Error rate |

**Distributed clustering**

Clustering is an unsupervised machine learning method that involves defining classes from data without knowing the labels of classes. In clustering, data is categorized into collections (or clusters) based on their similarities (52). Clustering algorithms apply when there is no class for prediction, so the instances divide into natural groups. Clustering distributed classifiers relies on the following: (i) A measure of classifier distance, (ii) An efficient algorithm for computing this distance measure for classifiers induced in physically distributed databases, and (iii) A clustering algorithm (23). The distributed clustering algorithms of consensus-based algorithm and distributed k-means algorithm are discussed in the following. A summary of these algorithms can be found in Table 2.

***Consensus-based algorithm***

Consensus clustering is a technique in which multiple clusters combine into a more stable single cluster that is better than the input clusters. It yields a stable and robust final clustering in agreement with multiple clusterings. Consensus clustering is a more robust approach that relies on multiple iterations of the chosen clustering method on sub-samples of the dataset (53). Vendramin et al. (54) presented a consensus-based algorithm for distributed fuzzy clustering that allows an automatic estimation of the number of clusters by using a distributed version of the Xie-Beni validity criterion.

***Distributed k-means***

K-means clustering is one of the most popular clustering algorithms due to its many advantages, such as simple mathematical concepts, quick convergence, and ease of implementation (55). K-means is an iterative process in which k centroids are determined. Then, each sample is assigned to the closest current centroid (assignment phases). The new centroid will be determined by the average of all samples in the same partition (refinement phase) (56).

Patel et al. (57) presented a parallel version of k-means focusing on privacy preservation. In distributed environments, where data mining becomes a collaborative effort, it is crucial to maintain privacy. The basic concept involves the use of a secret sharing mechanism to share information privately along with a code-based zero-knowledge identification scheme to add protection against malicious adversaries. Oliva et al. (58) suggested a fully distributed execution of the k-means clustering algorithm. It was applied for wireless sensor networks where each agent was provided with a high-dimensional observation. To spread information on current centroids across the network, the proposed algorithm uses a maximum consensus algorithm. Each agent employs this information to select the nearest centroid, thus segmenting the network into communities. For the purpose of updating centroids, meta-information is gathered by combining max-consensus and average-consensus algorithms. The agents are able to update the centroids locally once such information has been gathered.

A distributed k-means method has proposed by Benchara and Youssfi (59). It integrates a parallel virtual distributed computing model with a low-cost communication mechanism. K-means is implemented as a distributed service using an asynchronous communication protocol based on Advanced Message Queuing Protocol (AMQP). Datta et al. (60) discussed a distributed k-means clustering, in which data and computing resources are distributed over a large peer-to-peer network. Using two algorithms, it approximates the result produced by the centralized k-means clustering algorithm. The first algorithm is intended to be used in a dynamic peer-to-peer network. It is capable of producing clusterings only through the use of "local" synchronization. In the second algorithm, peers are uniformly sampled, and analytical guarantees are provided about the accuracy of clustering on an Internet-based peer-to-peer system. Ding et al. (61) studied distributed k-means clustering, in which dimensions of the data are distributed across multiple computers.

**Table 2** Distributed clustering.

| Algorithm | Articles | Year | No of references | Simulation/ Dataset | Evaluation metrics |
|---|---|---|---|---|---|
| Consensus-based algorithm | (53) | 2016 | 35 | • Wireless sensor networks (WSNs) | • within-cluster sum of squares (WCSS)<br>• Iteration time |
| | (54) | 2011 | 15 | • Two data sites | • Xie-Beni (XB) fuzzy clustering validity index |
| Distributed k-means | (59) | 2021 | 25 | • MRI image segmentation | • Number of iterations |
| | (61) | 2016 | 39 | • YearPredictionMSD | • Communication costs |
| | (57) | 2013 | 33 | • Mammal's Milk<br>• River dataset<br>• Water treatment dataset | • Communication Overhead<br>• Computation Overhead |
| | (58) | 2013 | 42 | • Wireless sensor networks | • Time complexity<br>• Memory complexity |
| | (60) | 2008 | 38 | • P2P network | • Accuracy<br>• Scalability<br>• Communication |

**Distributed deep learning**

Deep learning is a type of machine learning process that uses interconnected nodes or neurons in a layered structure that resembles the human brain (62). Neural networks consist of many computation units, known as neurons, which are connected and form the neural network. The input neurons of the network are actuated by input parameters. The neurons in the following layer are activated by weighted connections from neurons in the previous layer. To provide the desired functionality, usually classification or regression, the neural network must determine the appropriate weight value for every connection (63).

To overcome the problem of training DNN models, which requires a large volume of data and computational resources, a variety of parallel and distributed methods have been proposed (64, 65). These methods can be divided into four categories: data parallelism, model parallelism, pipeline parallelism, and hybrid parallelism (66). An overview of these algorithms is presented in Table 3.

***Data parallelism***

Data parallelism is a popular method for training a neural network that involves sharing a large-scale DNN among all computational workers (67, 68). In data parallelism, data samples are partitioned into mini-batches (Fig. 2) (69). During the computation of gradients, each node or worker contains one of the mini-batches, a replica of the neural network model, and independently computes gradients (usually using the Mini-Batch SGD) (70). The following steps are involved in training: i) Computation of local gradients by each worker; ii) Calculation of the new parameters of the DNN by combining all sub-gradients. iii) Distribution of the new parameters among the workers, and retraining of the DNN (71). To aggregate and update gradients, either a centralized architecture such as parameter server architecture (72), or a decentralized architecture such as All-Reduce (73) is used.

Data parallelization allows for processing large datasets that cannot be stored on a single machine and can increase the system's throughput through distributed parallel computing. However, data parallelism also has some challenges, including the overhead of parameter synchronization, optimization algorithms, and hardware limitations when the DNN model size is too large (64, 74).

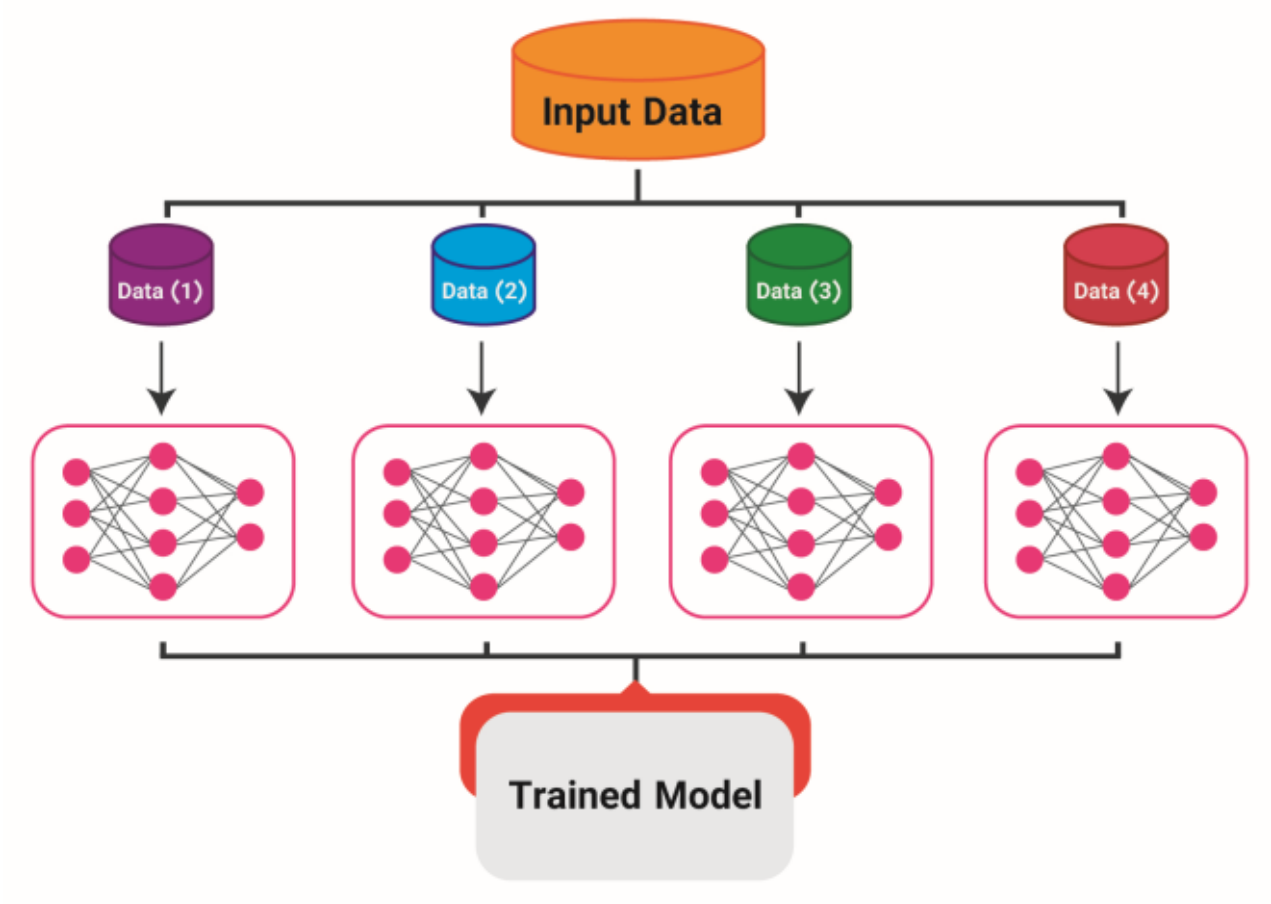

**Fig. 2** Data parallelism.

Dean et al. (67) introduced DistBelief, a framework for parallel distributed training of deep networks, and two new algorithms for large-scale distributed training: Downpour SGD and Sandblaster L-BFGS. Downpour SGD is an asynchronous variant of stochastic gradient descent (SGD), which is effective for training nonconvex deep learning models. As a distributed version of L-BFGS, Sandblaster L-BFGS makes efficient use of network bandwidth to train a single model on a large number of concurrent cores.

Zhang et al. (68) developed an algorithm for optimizing deep learning under communication constraints in a parallel environment. In this algorithm, elastic force is used to link the parameters calculated by local workers to the central variable stored by the parameter server. As a result, the amount of communication between the master and the local workers is reduced. Asynchronous and synchronous variants of this algorithm are available.

An algorithm for distributed SGD based on a communication trigger mechanism has been proposed by George et al. (69). They presented the Distributed Event-Triggered Stochastic GRAdient Descent (DETSGRAD) algorithm, which allows networked agents to update model parameters periodically in order to solve non-convex optimizations. The evaluation was conducted using MNIST with 60000 images for training and 10000 images for testing. During training, each agent used LeNet-5. There are two types of DETSGRAD: DETSGRAD-r, in which agents are randomly selected from the entire training set, and DETSGRAD-s, in which each agent has access to the images of only one class. According to the results obtained after 40 epochs with 10 agents, the accuracy of DETSGRAD-r and DETSGRAD-s was 98.33 and 98.51, almost similar to the accuracy of SGD-r and SGD-s with 98.97 and 98.87. Based on the results, it appears that DETSGRAD reduced inter-agent communication while maintaining similar performance.

Kim et al. (70) proposed Parallax, a framework that integrates parameter server and AllReduce architectures in order to optimize parallel data training by exploiting model parameter sparsity. ResNet-50 and Inception-v3 were used to classify images from the ImageNet dataset. In NLP, the LM model was trained using the One Billion Word Benchmark, and the NMT model was trained using the WMT English-German dataset. Image classification models are trained at the same speed as Horovod and 1.53x faster than TensorFlow. For NLP models, Parallax has achieved speedups of 2.8x and 6.02x compared to TensorFlow and Horovod.

The Dynamic Batch Size (DBS) strategy has been proposed by Ye et al. (71) for the distributed training of DNNs. According to the performance of previous epochs, DBS evaluates the performance of each worker, and then dynamically adjusts the batch size and dataset partition. DBS aims to optimize cluster utilization based on worker performance and can be used with all synchronous methods. The estimated batch size and dataset partition are employed in the next training. As compared synchronous stochastic gradient descent (S-SGD), DBS saved approximately 12% of the consumed time of each epoch on a scale of 4 and 10% on a scale of 8. The decreases can be attributed to cluster synchronization and communication costs, which are higher as the cluster expands.

Using data parallelism, Dong et al. (75) proposed a technique called "natural compression" that is an effective method for compressing data. It is based on the randomized rounding to the nearest (negative or positive) power of two, which can be computed in a "natural" manner without taking into account the mantissa. The natural compression method reduced the training time for ResNet110 by 26% (compared to only a 9% decrease for QSGD for the same setup) and 66% for AlexNet, compared to using no compression. In their study, they also presented convergence theory for distributed SGD to apply bidirectional compression at both the master and worker levels.

***Model parallelism***

Model parallelism is a technique used to speed up the training of DNNs by dividing a large model among multiple nodes or workers (Fig. 3) (76). Each node is responsible for part of the computation of model parameters, such as weights (74). However, the major challenges are how to break the model into partitions, as each model has its own characteristics, and the allocation of partitions to GPUs to maximize the efficiency of training (77). Furthermore, model parallelism alone is not scalable (78) due to high communication latency between devices.

A fully decoupled training scheme was proposed by Zhang et al. (79). A neural network was broken down into several modules (K) and trained on multiple devices. In the WRN-28-10 (CIFAR-10) case, delayed gradients slightly outperformed the decoupled greedy learning and achieved a speedup of 1.88x for K=2, 2.72x for K=3, and 3.20x for K=4. For the ResNet-101 (ImageNet) case, the delayed gradients achieved a 1.68x speedup for K = 2, and 2.1x and 2.3x for K = 3 and K = 4.

Huo et al. (80) proposed a Decoupled Parallel Back-propagation (DDG) algorithm for training feedforward neural networks. This algorithm splits the model and stores the delayed error gradient to solve the backward-locking problem. By increasing the number of GPUs from two to four, the method is able to reduce the total computation time by about 30% to 50%.

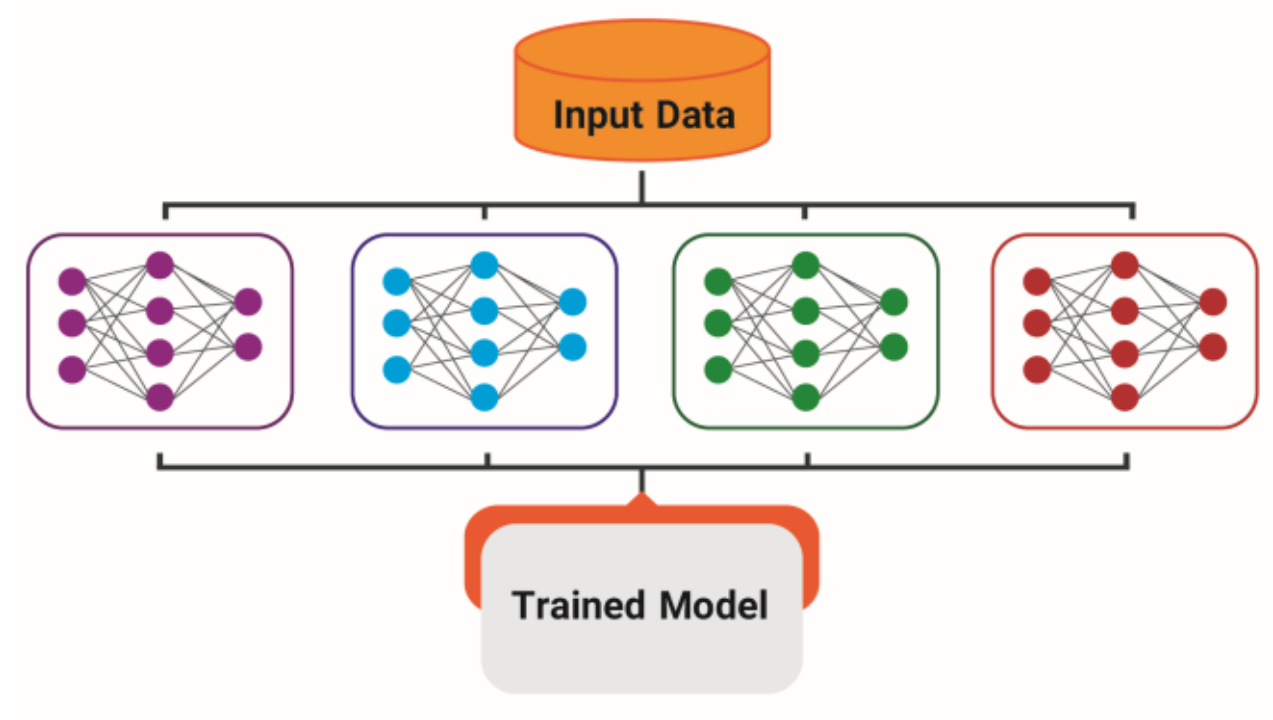

**Fig. 3** Model parallelism.

***Pipelining parallelism***

Pipeline parallelism is a technique used to divide training tasks for DNN models into sequential processing stages, and the results of each sequence are passed on to the next (81). Narayanan et al. (82) proposed PipeDream, a system that adds inter-batch pipelining to intra-batch parallelism to further improve parallel training throughput, allowing computation and communication to overlap more effectively and reduce communication. PipeDream updates model parameters for numerically correct gradient computations. In addition, forward and backward passes are scheduled concurrently on separate workers in order to minimize pipeline stalls. Furthermore, it automatically distributes DNN layers among workers so that work can be balanced and communication can be minimized. Compared to common intra-batch parallelism techniques, PipeDream performed 5.3 times faster.

Chen et al. (83) achived robust training accuracy by implementing a pipelined model and using a novel weight prediction technique. On a four-GPU platform, this method achieves an 8.91x speedup compared with data parallelism. Lee et al. (84) implemented a thread in each computer server to overlap computation and communication problems during model training. They achieved speedups of 62.97x and 77.97x for training VGG-A model on ImageNet. In parallel pipelines, there are two major problems: the slowest stage becomes a bottleneck. and scalability is limited (64).

***Hybrid parallelization***

Hybrid parallelization is a technique employed to minimize the communication overhead of DNN training by combining data and model parallelization techniques (85). Ono et al. (86) proposed a hybrid approach, which applies a model parallel to the RNN encoder-decoder in the Seq2Seq model, and data parallel to the attention-softmax. According to the results, using four GPUs increased training speed by 4.13 to 4.20 times over using one GPU alone. The solution proposed by Song et al. (87), HYPAR, partitions the feature map tensors (inputs and outputs), kernel tensors, gradient tensors, and error tensors among the DNN accelerators. During training, the goal of optimization is to search for a partition that minimizes the amount of communication. In an evaluation of classic Lenet to large-size model VGGs, HYPAR outperformed model parallelism and data parallelism alone. Compared to data parallelism, results showed a performance of 3.39x and an energy efficiency of 1.51x was achieved.

A hybrid parallelization method for training DNNs was proposed by Akintoye et al. (64), as well as a Genetic Algorithm Based Heuristic Resources Allocation (GABRA) approach for optimal partitioning on GPUs to maximize computing performance. Model parallelization includes neural network model partitioning and the GABRA mechanism. Asynchronous Stochastic Gradient Descent (ASGD) and ring All-Reduce mechanisms are used for data parallelization. The proposed approach that was applied to a 3D Residual Attention Deep Neural Network (3D-ResAttNet) using the ANDI dataset, achieved a 20% average improvement over existing parallel methods in terms of training time while maintaining accuracy.

The Heterogeneous Parameter Server (HeterPS) was proposed by Liu et al. (88) to facilitate the training of large-scale models using elastic heterogeneous computing resources. HeterPS consists of three modules: (i) Scheduling module for the DNN layer that generates a scheduling plan as well as a provisioning plan. In the provisioning plan, the number of computing resources of each type is defined, whereas in the scheduling plan, each layer is assigned the appropriate type of computing resource. (ii) A data management module that facilitates the transfer of data between clusters or servers. (iii) A distributed training module that exploits the combination of data parallelism and pipeline parallelism in order to parallelize the training process of the model. Experimental results indicated that the provisioning method can outperform baseline methods by up to 57.9% and the scheduling method can outperform state-of-the-art methods by up to 312.3% (monetary cost). Additionally, the framework has a throughput 14.5 times greater than TensorFlow.

Yadan et al. (89) achieved a speed improvement of 2.2x when training a large deep CNN using hybrid parallelization. Krizhevsky et al. (90) used hybrid parallelization to train a deep CNN and evaluated its performance by classifying 1.2 million images in the ImageNet LSVRC-2010 dataset. By combining parallel data and model training, Oyama et al. (91) increased throughput and minimized I/O scaling bottlenecks for a 3D CNN.

To address the challenge of aggregating sub-gradients effectively, several synchronous strategies have been used, including Parallel S-SGD (92, 93) and Bulk Synchronous Parallel (BSP) (94) among others.

**Table 3** Distributed deep learning.

| **Algorithm** | **Articles** | **Year** | **No of references** | **Simulation/ Dataset** | **Evaluation metrics** |
|---|---|---|---|---|---|
| Data parallelism | (75) | 2022 | 61 | • ResNet110 and AlexNet models on CIFAR10 | • Train loss<br>• Test Aacuracy |
| | (72) | 2022 | 24 | • Matrix Classification<br>• MovieLens Avazu-CTR | • Convergence Time per epoch<br>• Disk I/O<br>• Network communication |
| | (65) | 2021 | 138 | • ResNet-50 on ImageNet dataset<br>• ALBERT-large on WikiText-103 dataset | • Training time |
| | (71) | 2020 | 37 | • ResNet101 on CIFAR10 dataset | • Convergence<br>• Robustness |
| | (69) | 2019 | 53 | • LeNet-5 on MNIST dataset | • Accuracy |
| | (70) | 2019 | 46 | • ResNet-50 and Inception-v3 on ImageNet<br>• LM model on One Billion Word Benchmark<br>• NMT model on WMT English-German dataset | • Validation error<br>• Test perplexity<br>• BLEU |
| | (73) | 2018 | 20 | • Inception V3<br>• ResNet-101<br>• VGG-16 | • Images processed per second |
| | (68) | 2015 | 31 | • CNN on CIFAR and ImageNet datasets | • Test loss<br>• Test error |
| | (67) | 2012 | 29 | • ImageNet | • Accuracy |
| Model parallelism | (77) | 2021 | 72 | • GNN model on OGB-Product, OGB-Paper, UK-2006-05, UK-Union, Facebook datasets | • ROC |
| | (79) | 2021 | 29 | • ResNet and WRN models on CIFAR-10 dataset<br>• ResNet-18 and MobileNet v2 on Tiny-ImageNet | • Error rate |
| | (76) | 2019 | 30 | • AlexNet, Inception-v3 and ResNet-101 on ImageNet dataset<br>• RNNTC on Movie Reviews dataset<br>• RNNLM on Penn Treebank dataset<br>• NMT on WMT English-German dataser | • Accuracy |
| | (80) | 2018 | 25 | • ResNet on CIFAR | • Accuracy |
| Pipelining parallelism | (81) | 2020 | 29 | • AmoebaNet-D<br>• U-Net | • Throughput<br>• Speed up |
| | (82) | 2019 | 57 | • VGG-16 and ResNet-50 on ImageNet<br>• AlexNet on Synthetic Data<br>• GNMT-16 and GNMT-8 on WMT16 EN-De<br>• AWD LM on Penn Treebank<br>• S2VT on MSVD | • Accuracy<br>• Speed up |
| | (83) | 2018 | 50 | • VGG16, ResNet-152, Inception v4 and SNN on CIFAR-10<br>• Transformer on IMDb Movie Review Sentiment Dataset<br>• Residual LSTM on IMDb Dataset | • Speed up |
| | (84) | 2017 | 25 | • VGG-A model on ImageNet | • Speed up |
| Hybrid parallelization | (88) | 2023 | 64 | • MATCHNET, CTRDNN, 2EMB and NCE models | • Scheduling performance |
| | | | | | • Throughput |
| | (64) | 2022 | 57 | • 3D-ResAttNet on Alzheimer's Disease Neuroimaging Initiative (ADNI) database | • Speedup<br>• Accuracy<br>• Training time<br>• |
| | (91) | 2020 | 64 | • CosmoFlow and 3D UNet models | • MSE |
| | (86) | 2019 | 23 | • Seq2Seq RNN MT with attention on WMT14 and WMT17 datasets | • BLEU scores |
| | (87) | 2019 | 120 | • SFC, SCONV, Lenet-c, Cifar-c, AlexNet,VGG-A, VGG-B, VGG-C, VGG-D and VGG-E models on MNIST, CIFAR-10 and ImageNet datasets | • Energy efficiency<br>• Performance<br>• Total communication |
| | (85) | 2018 | 67 | • AlexNet and VGG models | • Communication Overhead<br>• Training time<br>• Speed up |
| | (90) | 2017 | 33 | • CNN on ImageNet LSVRC-2010 dataset | • Error rate |
| | (89) | 2013 | 12 | • ImageNet dataset | • Error rate |

**Distributed deep reinforcement learning**

Reinforcement learning is a learning algorithm that involves learning by interacting with the environment through actions, observations, and rewards. Reinforcement learning faces a major challenge when it comes to learning good representations of high-dimensional states or action spaces (95). DRL combines reinforcement learning with deep learning, allowing the representation of a continuous state or action, which was difficult for a table representation (96). However, DRL faces technical and scientific challenges such as data inefficiency, multi-task learning, and exploration-exploitation trade-offs. To overcome these challenges, distributed DRL was introduced. In distributed DRL, agents can run simultaneously on several computers allowing for parallelization of the learning process (97).

Nair et al. (98) introduced the GORILA (General Reinforcement Learning Architecture), which is similar to DQN (99), but with multiple workers and learners, and the SGD is computed using the DistBelief (67) method. In the GORILA architecture, there are N different actor processes, which are applied to N corresponding instances of the same environment. The Q-network is replicated in each actor, which determines its behavior. A parameter server periodically synchronizes the parameters of the Q-network. There are N learner processes in GORILA. Learners contain replicas of the Q-network and are responsible for computing desired changes to its parameters. There are many ways in which a reinforcement learning agent may be parallelized using the GORILA architecture. One approach is parallel acting, where large quantities of data can be generated and then processed by a single serial learner using a global replay database. Alternatively, a single actor can generate data into a local replay memory, after which multiple learners can process this data in parallel to maximize the effectiveness of their learning. Using the Arcade Learning Environment, GORILA was evaluated on 49 Atari 2600 games. In 25 games, GORILA achieved 75% of the human score or higher.

The A3C (Asynchronous Advantage Actor-Critic) algorithm was proposed by Mnih et al. (100), in which multiple agents generate data in parallel asynchronously, and DNN controllers are optimized through gradient descent asynchronously. Similar to the GORILA, actor-learners were used asynchronously, however, instead of using separate machines and a parameter server, multiple CPU threads were used on a single computer. The cost of communication can be eliminated by keeping learners on the same computer. It is likely that multiple actors and learners explore different aspects of the environment simultaneously. This approach can maximize diversity by utilizing different exploration policies for each actor-learner. As multiple actor-learners use online updates in parallel, the overall changes being made to the parameters are more likely to be less correlated over time than the changes made by a single agent. As a result of parallelization, the data is also diverse and decorated, which provides a more practical alternative to experience replay.

IMPALA (Importance Weighted Actor-Learner Architecture) is a scalable DDR learning algorithm proposed by Espeholt et al. (101). IMPALA uses GPUs and distributed deep learning methods to update mini-batches of data in parallel, which allows it to train large neural networks efficiently with a distributed set of learners uses synchronized parameter updating and is capable of training. In IMPALA, there is a distribution of parameters across the learners and actors retrieve the parameters from all the learners simultaneously while only sending observations to one learner. IMPALA outperforms A3C-based agents on DMLab-30, achieving a 49.4% vs. 23.8% human normalized score.

Heess et al. (102) introduced Distributed Proximal Policy Optimization (DPPO), a DRL approach based on the principle of proximal policy optimization (PPO) (103). In DPPO, the collection of data and the calculation of gradients are distributed among the workers. The experiments have been conducted with both synchronous and asynchronous updates, and the results have shown that averaging gradients and applying them synchronously leads to better results.

Ape-X (104) is a distributed architecture for DRL that decouples acting from learning. In Ape-X, a shared neural network is used to select actions by actors and the resulting experience is stored in a shared experience replay memory. The neural network is updated by replaying samples of experience, with prioritizing given to the most significant data generated by the actors.

SEED (Scalable, Efficient Deep-RL) was proposed by Espeholt et al. (105). SEED RL utilizes modern accelerators to improve the speed and efficiency of DRL. SEED RL uses three types of threads: the inference thread, data prefetching threads, and training threads. The inference thread receives a batch of observations, rewards, and episode termination flags, while data prefetching threads sample data as it is added to a FIFO queue or replay buffer. For each of the TPU cores participating in training, the trajectories are pushed to a device buffer. In comparison with the baseline IMPALA, SEED improves the speed by 1.6x (with 2 cores), 4.1x (8 cores), and 4.5x (if the batch size is increased linearly with 5 cores).

Acme (106) is a research framework that helps with algorithm development. Acme aims to increase reproducibility in reinforcement learning and simplify the development of novel and creative algorithms. Acme's main advantage is that it can be used to implement large-scale distributed reinforcement learning algorithms enabling operation at enormous scales while maintaining the inherent readability of the code. In most cases, algorithms implemented with Acme result in a distributed agent with a number of separate (parallel) acting, learning, diagnostic, and helper processes. Acme's main design decision, however, is to reuse the same components across simple, single-process implementations and large-scale distributed systems.

In a recent paper, Dai et al. (107) proposed a "hybrid near-on policy" DRL framework, called Coknight, which leverages a game theory-based DNN partition approach to achieve fast and dynamic partitioning in distributed DRL architectures.

Table 4 provides an overview of these algorithms.

**Table 4** Distributer DRL.

| **Algorithm** | **Articles** | **Year** | **No of references** | **Simulation/ Dataset** | **Evaluation metrics** |
|---|---|---|---|---|---|
| Distributed deep reinforcement learning | (107) | 2022 | 42 | • Atari games | • Onvergence rate<br>• Convergence time<br>• Running time<br>• GPU usage<br>• Memory usage<br>• Bandwidth consumption |
| | (106) | 2020 | 127 | • 5 Atari games: Asterix,<br>• Breakout, MsPacman, Pong and SpaceInvaders<br>• Arcade Learning Environment | • Mean and standard deviation<br>• Speed |
| | | | | • DeepMind Control suite<br>• Gym environments | |
| | (105) | 2019 | 53 | • Atari-57<br>• DeepMind Lab<br>• Google Research Football | • Training cost<br>• Speed |
| | (101) | 2018 | 41 | • Atari-57<br>• DMLab-30 | • Median and Mean Human-Normalized scores |
| | (104) | 2018 | 40 | • Atari games | • Median and Mean Human-Normalized scores |
| | (100) | 2016 | 43 | • Atari games<br>• TORCS 3D<br>• Mujoco<br>• Labyrinth | • Median and Mean Human-Normalized scores |
| | (98) | 2015 | 19 | • 49 games from Atari 2600 games | • Human Score |

## Conclusions and Research Directions

Distributed machine learning is becoming increasingly important due to the increase in data, the need for more accurate models, the ability to solve complex problems, and the reduction of computation time.

Researchers have proposed different methods for distributing machine learning algorithms, including distributed algorithms for classification, clustering, deep learning, and reinforcement learning. In the case of traditional machine learning methods (clustering and classification algorithms), some studies have attempted to develop distributed versions of them. Our study reviewed the distribution of Boosting, SVM, Consensus-based, and K-means algorithms. For deep learning, there are four types of parallelism: data parallelism, model parallelism, pipelining parallelism, and hybrid parallelism. The majority of these studies considered neural networks such as ResNet, VGG, and AlexNet. In the case of reinforcement learning, researchers have proposed various distributed reinforcement learning algorithms, including A3C, IMPALA, DPPO, Ape-X, SEED RL, and Acme. Distributed machine learning has several limitations that need to be addressed in future research. These limitations include:

- Lack of attention to distributed traditional machine learning: There has been a significant focus on distributed deep learning in recent studies and less attention has been paid to distributed traditional machine learning. Although machine learning algorithms have their advantages and have shown promising results in a number of areas, they have not been studied as extensively as deep learning in distributed systems.

- Lack of benchmarks: Most studies used MNIST and ImageNet datasets to evaluate their proposed method, but there is no benchmarking to evaluate and compare the performance of existing approaches. Researchers considered a wide range of models, datasets, and evaluation metrics, and even in distributed RL, each study evaluated its method on different types of Atari games. Consequently, benchmarks are necessary to compare the results of different methods.
- Interpretability: Even though DNNs have excellent performance in many areas, understanding their results, particularly in distributed systems, can be challenging. A model's interpretability can help to provide insight into the relationship between input data and the trained model, which is particularly useful in critical domains like healthcare. The interpretability of distributed algorithms remains an open problem.
- New issues: New subjects arise when we try to have distributed algorithms, including the way data and model are partitioned, optimality, delay of the slowest node, communication overhead, scalability, and aggregation of results. These issues need to be addressed to succeed at distributed training and to make it more accessible to data scientists and researchers.

Therefore, this is an open line of research that will have to overcome these new challenges in the future.

**Table 5** List of abbreviations.

| | |
|---|---|
| AI | Artificial intelligence |
| BSP | Bulk Synchronous Parallel |
| DNN | Deep Neural Network |
| DRL | Deep Reinforcement Learning |
| IOT | Internet of Things |
| ML | Machine Learning |
| NLP | Natural Language Processing |
| PPO | Proximal Policy Optimization |
| SVM | Support Vector Machines |
| SGD | Stochastic Gradient Descent |
| S-SGD | Synchronous Stochastic Gradient Descent |
| WSNs | Wireless Sensor Networks |